\documentclass[11pt]{article}
\usepackage{nodalida2023}
\usepackage{times}
\usepackage{url}
\usepackage{latexsym}
\aclfinalcopy

\usepackage[utf8]{inputenc}

\usepackage{amsmath}
\usepackage{booktabs}
\usepackage{graphicx}
\usepackage{enumitem}

\title{The Nordic Pile: A 1.2TB Nordic Dataset for Language Modeling}

\author{Joey Öhman, Severine Verlinden, Ariel Ekgren,$^\star$ Amaru Cuba Gyllensten, Tim Isbister, \\ {\bf Magnus Sahlgren} \\
  AI Sweden \\
  $^\star$Corresponding author: \texttt{ariel.ekgren@ai.se} \\\AND
  Evangelia Gogoulou, Fredrik Carlsson \\
  RISE \\}

\begin{document}

\maketitle
\begin{abstract}
Pre-training Large Language Models (LLMs) require massive amounts of text data, and the performance of the LLMs typically correlates with the scale and quality of the datasets.
This means that it may be challenging to build LLMs for smaller languages such as Nordic ones, where the availability of text corpora is limited. In order to facilitate the development of the LLMS in the Nordic languages, we curate a high-quality dataset consisting of 1.2TB of text, in all of the major North Germanic languages (Danish, Icelandic, Norwegian, and Swedish), as well as some high-quality English data. This paper details our considerations and processes for collecting, cleaning, and filtering the dataset. 
\end{abstract}

\section{Introduction}

Recent work on Large Language Models (LLMs) show that for current architectures, model performance is strongly correlated with its scale \citep{gpt3_2020, gopher_2021, gptneox_2022, bloom_model_2022, palm_2022}, with some capabilities emerging only past a certain parameter count. However, with the constantly growing model architectures, the optimal dataset size and required to compute increases in a proportional manner \citep{scaling_2022, chinchilla_2022}. This suggests that access to a massive dataset is a fundamental requirement for pre-training such LLMs. For lower-resourced languages, this may be a limiting factor, which often leads practitioners to rely on English or multilingual models.

For the languages in the North Germanic language group, there is limited open access to large textual datasets. This may be a limiting factor for the development of LLMs for these languages. We therefore create a training dataset consisting of 1.2TB of high-quality multilingual texts in Danish, English, Icelandic, Norwegian, and Swedish. We refer to the dataset as {\bf The Nordic Pile}, since the collection process has been inspired by its English predecessor The Pile \cite{pile_2021}. This paper details our considerations and processes for collecting, cleaning, and filtering the dataset in order to build a well-performing Nordic LLM. 

\begin{figure*}[]
\includegraphics[width=\linewidth]{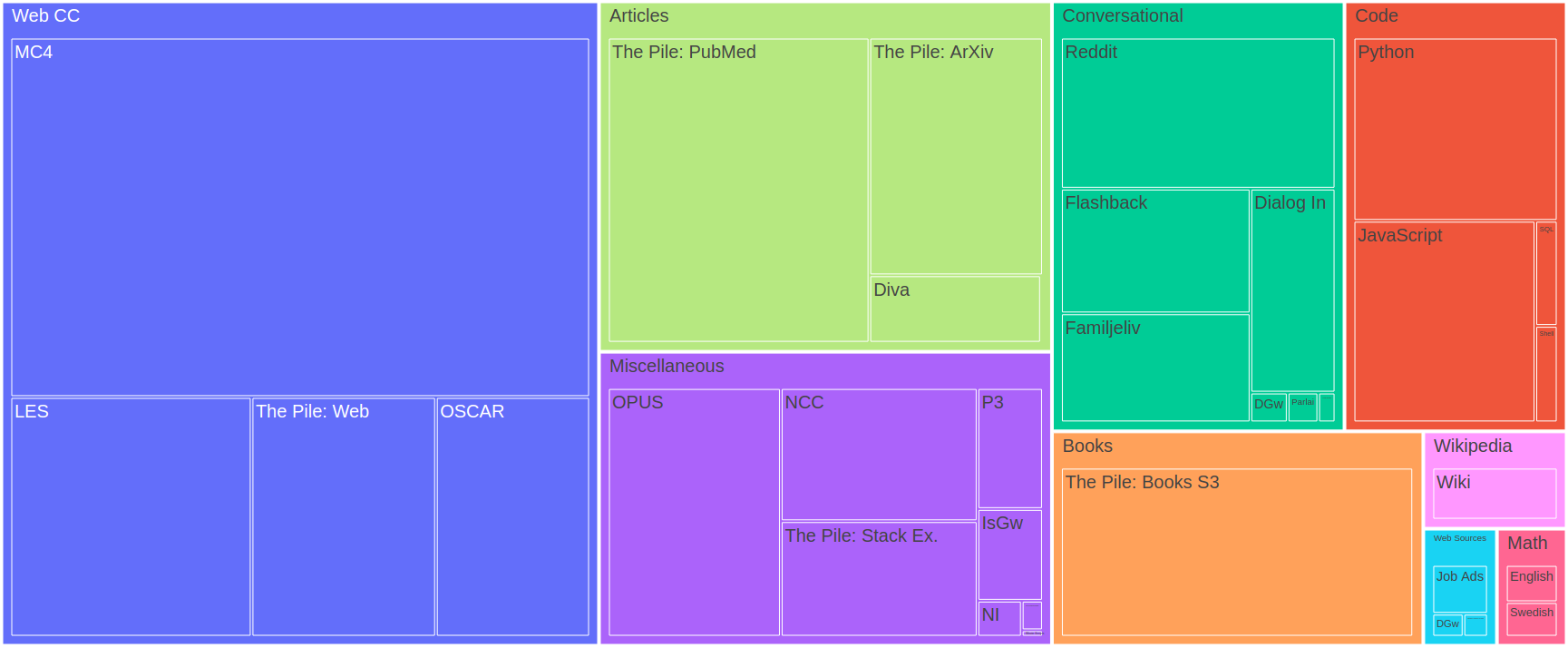}
\centering
\caption{Tree map visualizing our final dataset.}
\label{fig:data_tree_map}
\end{figure*}

\section{Related Work}

The rapid progress within the field of Natural Language Processing (NLP), has to a great extent benefited from having access to open datasets. These are often derived from Common Crawl\footnote{\url{commoncrawl.org/the-data/}}, such as C4 \citep{t5_2020}, OSCAR \citep{oscar_2019,oscar_2020} and The Pile \citep{pile_2021}. While the majority of these datasets have included some form of quality filtering and deduplication, recent work has applied more aggressive filters and deduplication processes to their datasets prior to LLM pre-training.


\citet{ccnet_2020} propose an automatic pipeline for extracting massive high-quality monolingual datasets from Common Crawl. The pipeline includes deduplication and language identification using fastText \citep{joulin2016bag}, and is extended by removing low-quality documents, using a small Kneser-Ney language model trained on Wikipedia. The authors show that their filtering process results in better model performance.

\citet{gpt3_2020} present GPT-3 and describes their corpus, which is a filtered and deduplicated version of Common Crawl, extended with several curated high-quality datasets. \citet{gopher_2021} train Gopher on their cleaned corpus \textit{MassiveText}, a diverse dataset with a large portion derived from web data. Their dataset pipeline consists of six stages: content filtering, text extraction, quality filtering, repetition removal, document deduplication, and test-set filtering. \citet{bloom_data_2022} also perform thorough quality filtering to create their 1.6TB multilingual ROOTS dataset. 

\citet{deduplicating_2022} extensively experiment with two different deduplication algorithms, MinHash Locality Sensitive Hashing (MinHash LSH) and Exact Substring Matching, and evaluate their effects by training LLMs, showing that deduplication indeed improves the model performance.

Our data pipeline is heavily inspired by the related literature, and parts of the above work are hence revisited and further explained in Section \ref{sec:data_pipeline}.


\section{Data Collection}

When collecting language modeling datasets, both quality and quantity are vital. The largest language models are often pre-trained on more than 1TB of text, with higher quality typically yielding better-performing models. In order to collect these massive amounts of data, manual methods are infeasible. Therefore, The Nordic Pile is composed mostly of existing sources, with a large portion of these originating from derivatives of Common Crawl data, such as OSCAR \citep{oscar_2019,oscar_2020} and Multilingual C4 (mC4) \citep{mt5_mc4_2021}, which is a language-filtered version of C4 \citep{t5_2020}.

In order to ensure the versatility of models trained on this dataset, we strive for a high degree of diversity in the data. This entails collecting data not only from Common Crawl, but also from other categories of data, such as conversational forums, academic articles, books, code, task-related datasets, and more. While most of the data originate from public pre-curated datasets, parts of The Nordic Pile were scraped using the Trafilatura library \citep{trafilatura_2021} or curated from other sources. For instance, after downloading the Reddit data from Pushshift \citep{pushshift_2020}, we constructed conversational trees in order to enable sampling of sequential conversations. Other examples involve building and translating templates for language-parallel sentences from OPUS\footnote{\url{opus.nlpl.eu/index.php}} \citep{opus_2004} or generated mathematical tasks\footnote{\url{github.com/deepmind/mathematics\textunderscore dataset}} \citep{math_2019}. 

Our data sources can be divided into the following nine categories, based on their domains and quality, illustrated in Figure \ref{fig:data_tree_map}:

\begin{itemize}[topsep=0pt,itemsep=-1ex,partopsep=1ex,parsep=1ex]
\item \textbf{Articles:} Academic papers.
\item \textbf{Books:} High-quality books, e.g. fiction, novels.
\item \textbf{Code:} Code in the programming languages \texttt{Bash}, \texttt{JavaScript}, \texttt{Python}, and \texttt{SQL}.
\item \textbf{Conversational:} Primarily social forums, such as Reddit.
\item \textbf{Math:} Mathematical problems and solutions.
\item \textbf{Miscellaneous:} Data which do not belong in any other category, often task-specific, such as parallel data.
\item \textbf{Web CC:} Web data derived from Common Crawl.
\item \textbf{Web Sources:} Web data from other sources, e.g. scraping.
\item \textbf{Wikipedia:} Official Wikipedia dumps.
\end{itemize}

A complete overview of our collected resources is available in Appendix \ref{sec:appendix_data_sources}. 

The inclusion of multiple datasets from the same domain inherently introduces overlaps between datasets, which is sub-optimal for efficient pre-training. Our method to address this issue is explained in Sections \ref{sec:exact_deduplication} and \ref{sec:fuzzy_deduplication}. All collected documents are formatted consistently in the \texttt{JSON Lines}\footnote{\url{jsonlines.org}} format as a preparation for further data processing.



\section{Data Processing Pipeline}
\label{sec:data_pipeline}

Our data processing pipeline consists of 7 steps: 
\begin{enumerate}[topsep=0pt,itemsep=-1ex,partopsep=1ex,parsep=1ex]
    \item normalization, 
    \item metrics, 
    \item quality filtering, 
    \item exact deduplication, 
    \item language segmentation, 
    \item fuzzy deduplication, 
    \item merging. 
\end{enumerate}
Each of these steps are described in more detail below. All of these steps, except for deduplication and merging, are done completely on the document-level, and are thus embarrassingly parallel. The source code for the first 3 steps\footnote{\url{https://github.com/SeverineVerlinden/data_analysis_base_pile}} and 4 last steps\footnote{\url{https://github.com/JoeyOhman/Megatron-deduplication}} is publicly available on GitHub.

\subsection{Normalization}

This step ensures that documents are consistently encoded, and perform the following operations:

\textbf{Non-printing character removal} removes unwanted characters that may be hidden in the document, for instance, control characters such as soft hyphens or non-breaking spaces.

\textbf{Whitespace normalization} converts any form of whitespace to a standard whitespace.

\textbf{Unicode normalization} of the text using NFC Unicode normalization\footnote{\url{unicode.org/reports/tr15/#Norm_Forms}}. The motivation to opt for NFC is its non-destructive transformation that maintains a consistent and informative format.

    
    

\subsection{Metrics}

After normalizing the text, document-level metrics are added as metadata. This is primarily done to satisfy prerequisites of the following stages of the data processing pipeline, as some of these are later required. Moreover, these metadata metrics assist in the overall data analysis. We add the following metrics to each document:

\begin{itemize}[topsep=0pt,itemsep=-1ex,partopsep=1ex,parsep=1ex]
    \item \textit{lang}: language identified using fastText \citep{fasttext_langid_2016_1, fasttext_langid_2016_2},
    \item \textit{num\_chars}: number of characters,
    \item \textit{num\_utf8bytes}: number of UTF-8 encoded bytes,
    \item \textit{num\_words}: number of words,
    \item \textit{num\_sents}: number of sentences,
    \item \textit{md5}: 128-bit MD5 hash as hexadecimal string.
\end{itemize}

\subsection{Quality Filtering}

Documents of poor quality not only negatively impact model performance but may also increase the risk of divergence and other problematic behaviors during LLM pre-training. To remedy this, we have aggressively filtered our data using many different filters. Most of these filters are inspired by the data processing work described in \textit{Gopher} \citep{gopher_2021} and \textit{ROOTS} \citep{bloom_data_2022}. Due to the nature of the different categories of data, we have not used all filters for all data categories, see Appendix \ref{sec:appendix_filter_configs} for details on where the different filters are used. Whenever a document does not meet a filter criterion, we save that information in the document's metadata to facilitate later data analysis. Furthermore, when a document is removed, it is not present in the succeeding pipeline stages. Below follows descriptions for each of the 16 filters used, if the statement about the document is false, the document is removed.

\textbf{Alpha Present:} at least $80\%$ of the words contain an alphabetic character.

\textbf{Blacklist URLs:} the URL is not malformed, and the domain, file extension, and URL are not blacklisted. If a URL is not available, this filter is skipped.

\textbf{Digit Fraction:} the digit to character ratio is less than $0.2$.

\textbf{Document Length:} \textit{num\_chars} is greater than 50.

\textbf{Ellipsis To Word Ratio:} the ellipsis to word ratio is less than $0.1$.

\textbf{Flagged Words:} each flagged word is coupled with a weight in the range $[0, 1]$, where higher weights are worse. The document contains less than $4$ total flagged words and less than $3$ unique flagged words, and the sum of weights for flagged words in the document is less than $num\_words / 100$.

\textbf{Hashtag To Word Ratio:} the hashtag to word ratio is less than $0.1$.

\textbf{Initial Bullet Point:} less than $90\%$ of the lines start with a bullet point, or such lines occur less than $3$ times.

\textbf{Mean Line Length:} let $MeanMed$ be the operation of computing the mean of the median value and mean value. $MeanMed$ number of characters per non-empty line is greater than $9$, and $MeanMed$ number of words per non-empty line is greater than or equal to $2.1$.

\textbf{Mean Word Length:} average word length is in the range $[2, 10]$.

\textbf{Repetitive BSP:} the document is not considered repetitive, according to the filter for repetition used for \textit{ROOTS} \citep{bloom_data_2022}. Details for this filter were found in their source code, and is, complementary to the \textit{Repetitive Gopher} filter, based on characters and words. 

\textbf{Repetitive Gopher:} the document is not considered repetitive, according to the Repetition Removal filter described in Gopher \cite{gopher_2021}, pages 40-41, based on n-grams, lines, and paragraphs. Our implementation differs in that it is word-based whereas the original implementation is token-based. To account for this, and our multilingual data, the thresholds are adjusted and are shown in Table \ref{tab:gopher_repetitive_thresholds}.

\begin{table}[]
\small
\begin{center}
\caption{Repetitive thresholds used in our implementation of the Repetitive Gopher filter.}
\label{tab:gopher_repetitive_thresholds}
\begin{tabular}{@{}lc@{}}
\toprule
Measurement                            & Threshold \\
\midrule
Duplicate line fraction                & 0.35      \\
Duplicate paragraph fraction           & 0.35      \\
Duplicate line character fraction      & 0.20      \\
Duplicate paragraph character fraction & 0.20      \\
Top 2-gram character fraction          & 0.25      \\
Top 3-gram character fraction          & 0.23      \\
Top 4-gram character fraction          & 0.21      \\
Duplicate 5-gram character fraction    & 0.20      \\
Duplicate 6-gram character fraction    & 0.19      \\
Duplicate 7-gram character fraction    & 0.18      \\
Duplicate 8-gram character fraction    & 0.17      \\
Duplicate 9-gram character fraction    & 0.16      \\
Duplicate 10-gram character fraction   & 0.15      \\
\bottomrule
\end{tabular}
\end{center}
\end{table}

\textbf{Stop Word} contains at least 2 stop words, and at least $10\%$ of all words are stop words. 

\textbf{Supported Language} \textit{lang} is one of Danish, English, Icelandic, Norwegian, or Swedish.

\textbf{Trailing Ellipsis} less than $30\%$ of the lines end with an ellipsis, or such lines occur less than $3$ times.

These filters complement each other, and are designed to capture different characteristics of low-quality documents. While all filters are not active for all data categories, the goal is that the subset of filters that are should capture the majority of non-desired documents. Some filters can be tweaked through parameters and could be adjusted for each language and data category similar to \citet{bloom_data_2022}. In our data pipeline, however, these are for simplicity fixed to values that seemed overall suitable through quantitative and qualitative evaluation. This is primarily due to limited resources and yields a less specialized yet robust filtering method. 

\subsection{Exact Deduplication}
\label{sec:exact_deduplication}


Even if the fuzzy deduplication would most likely also remove identical documents, we perform an initial exact deduplication step. The reason for this is primarily technical; fuzzy deduplication is computationally expensive and requires a lot of memory. Identifying exact duplicates is trivial and can ease the burden of fuzzy deduplication. 

We sequentially iterate through all documents and maintain a set of seen MD5 hashes. When a document with a previously seen MD5 hash is encountered, we simply mark it as a duplicate for later analysis and remove it. 

\subsection{Language Segmentation}

The sole purpose of this step is to prepare for fuzzy deduplication. Since fuzzy deduplication is computationally heavy, we separate the supported languages and fuzzily deduplicate them separately. Therefore, using the previously identified document language, documents are split into separate language-specific subsets on the document level. 

\subsection{Fuzzy Deduplication}
\label{sec:fuzzy_deduplication}

With the many subsets included in The Nordic Pile, some overlaps emerge, especially within the Common Crawl-based corpora. However, documents may be considered duplicates while not being identical. Consider an example where only a few characters or words differ, or where only a subset of the document is duplicated. These are not trivial to identify and decisions have to be made about how similar documents must be to be considered duplicates. There are two major algorithms used for this in the literature, evaluated and compared by \citet{deduplicating_2022}. These are Exact Substring Matching using Suffix Arrays \citep{suffixarray_1993} and MinHash LSH \citep{minhash_1997}. For explanations of how these two methods work and are implemented, along with selected parameters, see Appendix \ref{sec:appendix_deduplication_algorithms}.

We opted for the MinHash LSH solution, and its purpose is to identify groups of duplicated documents, from which only one is kept. Despite it being an efficient algorithm, deduplication of massive datasets is inevitably computationally expensive. During this work, we had access to machines with at most 256GB RAM. With our limited resources, it is difficult to deduplicate the entire dataset as one. Furthermore, experiments showed that our implementation required roughly three times the dataset size for deduplication. To avoid crashes, we added some margins and never deduplicated subsets of more than 80GB at a time. Therefore, we relied on sharding the data, and deduplicating shards separately (intra-shard) or pair-wise (inter-shard). 

The first form of sharding is a product of the previous Language Segmentation step, and we do not deduplicate anything cross-language. This should not have a major impact on the results, assuming that fuzzy duplicate documents are rarely identified with different languages. For all languages except Icelandic, the datasets were of sizes greater than 80GB, forcing us to further segment the data into smaller shards. We decided that our Nordic-language datasets were more prone to include many duplicate documents since we included several data sources from Common Crawl that may overlap for these languages. So, our English data was deduplicated intra-shard, and the Nordic data (except Icelandic) was deduplicated inter-shard.


\subsubsection{Intra-shard Deduplication}

Intra-shard deduplication divides the data of one language into shards, each maximally 80GB. Then, each shard is deduplicated separately, and duplicates across shards will not be identified. This type of shards is hence sub-optimal for finding duplicates, but requires significantly less resources. Since the English data was already to a large extent deduplicated with less overlapping subsets, we deemed this method sufficient. When sharding the English data, we aimed for having each shard include as similar data as possible.

\subsubsection{Inter-shard Deduplication}

Inter-shard deduplication is a proxy for complete deduplication and theoretically achieves the same result as deduplicating everything together. We divide the data of a language into shards of at maximum 40GB. This accounts for half of the previous memory constraint, to enable pair-wise deduplication of shards. Using this approach, we sacrifice computational performance to reduce the memory constraint of complete deduplication. After dividing the data into $N$ shards of accepted size, there are two steps to this approach:

\textbf{Pair-wise deduplication} deduplicates all shard pairs separately, each pair as one concatenated dataset. This setup is equivalent to a complete graph with $N$ nodes and $E$ edges, each node and edge corresponding to a shard and pair-wise deduplication respectively. The number of edges of a complete graph is shown in Equation \ref{eq:deduplication_pairs}. The computational sacrifice of this method is now evident; each shard is part of $N-1$ pairs and, therefore, redundantly deduplicated many times. While the stochastic nature of MinHash LSH may entail that more than one deduplication of a single shard is beneficial, it is hardly compute efficient. Nevertheless, this step outputs a set of groups containing duplicate documents, for each shard-pair, which are combined in the next step.

\begin{equation}
    \label{eq:deduplication_pairs}
    E = {N \choose 2} = \dfrac{N(N-1)}{2}
\end{equation}

\textbf{Merging of duplicate groups} combines the duplicate groups by identifying connected components. The naive approach would keep one document from each of the found groups, but would allow for significantly more duplicate documents to slip through.

Since we have deduplicated all shard pairs, each duplicate group may only contain documents from at most two shards, while a complete deduplication could find groups with documents from all shards. By merging groups from all shard pairs, we can approximate the behavior of a full deduplication. For instance, consider any groups $G_{ij}$ and $G_{jk}$, including documents from shards $S_i$, $S_j$, and $S_j$, $S_k$ respectively. With at least one common document in these groups, we can consider them part of the same group and hence merge them. This is equivalent to an undirected graph of document nodes, each node connected to the other nodes in its duplicate group. Additionally, whenever two groups share a document we merge these nodes and thus create a connection between the groups. This reduces the problem to identification of connected components, where only one document is kept from each component. 

We have through this process evaded the memory constraints, and achieved thorough fuzzy deduplication of our data in the Nordic languages. Appendix \ref{sec:shards} provides an overview of the selected sharding method for each language. While our solution satisfies the desiderata, there is ample room for improvement. Optimizations may include streaming data structures from disk, and specifically for inter-shard deduplication, part of the work required for each shard, e.g. computing fingerprints, could be reused for the succeeding deduplication. 

\subsection{Language \& Shard Merging}

This stage is theoretically trivial and was merely a technical hurdle given the data size and a large number of files. At this point in the pipeline, the data is segmented into two levels, language, and shards. Here we merge all non-removed documents to the original unified dataset, divided only into the original categories. Now, the dataset is in its final format, with only documents deemed clean and unique, along with additional document-level metadata.

While the absolute majority of the data was processed in the pipeline described, some deviations emerged for practical reasons. Examples of this are more data coming in after the deduplication step and additional filtering required after encountering errors or undesired behavior during tokenization or model training. We list these, mostly negligible, deviations in Appendix \ref{sec:appendix_deviations}.


\section{Results}

In this section, we present our insights from the analysis of our collected data. More specifically, we visualize the impact of our different pipeline steps on the data. In our data collection, we created a dataset with 1.5TB from which 1.2TB remained after the data pipeline. 



\begin{figure}[h!]
\includegraphics[width=\linewidth]{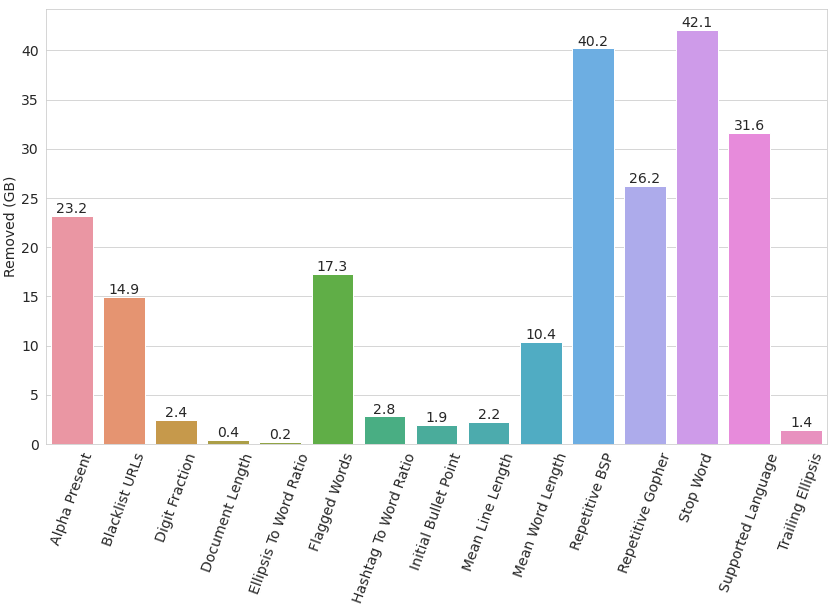}
\centering
\caption{The data size removed by each quality filter. Note that one document may be filtered by several filters.}
\label{fig:quality_filters_individual}
\end{figure}

Figure \ref{fig:quality_filters_individual} shows how much data was marked to be removed by each individual filter. While some filters overlap more than others, this provides a hint of which filters were most vital. A large portion of the removed documents seems to have been filtered via the \textit{Stop Word} and repetitive filters. 

\begin{figure}[h!]
\includegraphics[width=\linewidth]{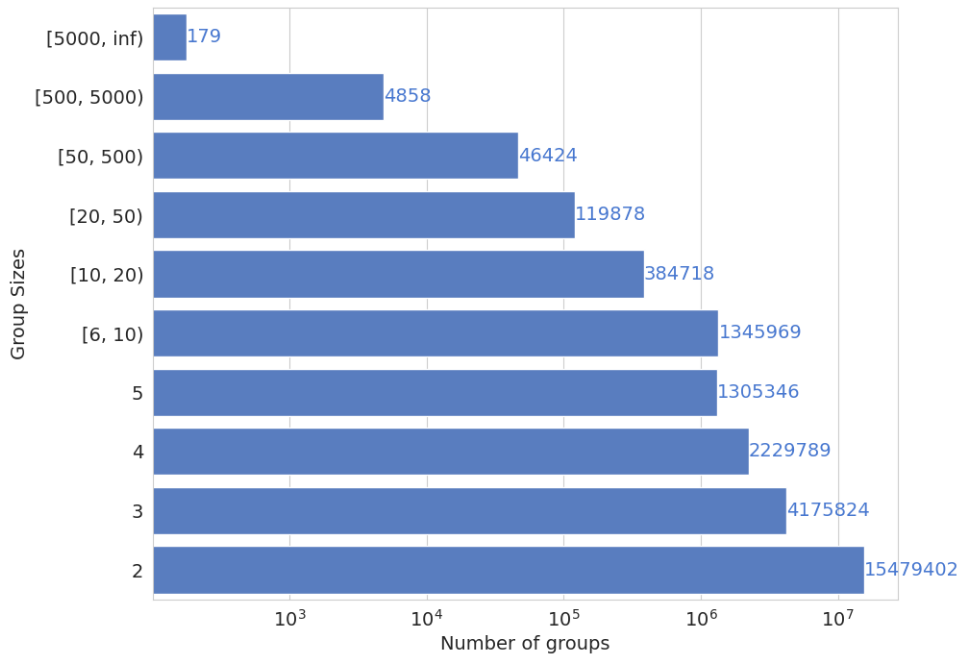}
\centering
\caption{Distribution over duplicate group sizes.}
\label{fig:deduplication_group_distribution}
\end{figure}

To gain an understanding of the impact of the fuzzy deduplication process, Figure \ref{fig:deduplication_group_distribution} depicts the distribution over duplicate group sizes. While there are surprisingly large groups present, the majority of fuzzy duplicate documents removed were part of smaller groups. Roughly 7\% of these duplicate documents originate from the generated mathematics datasets and compose some of the largest groups. For examples of documents removed and their corresponding group sizes, see Appendix \ref{sec:appendix_fuzzy_groups}.

\begin{figure}[h!]
\includegraphics[width=0.8\linewidth]{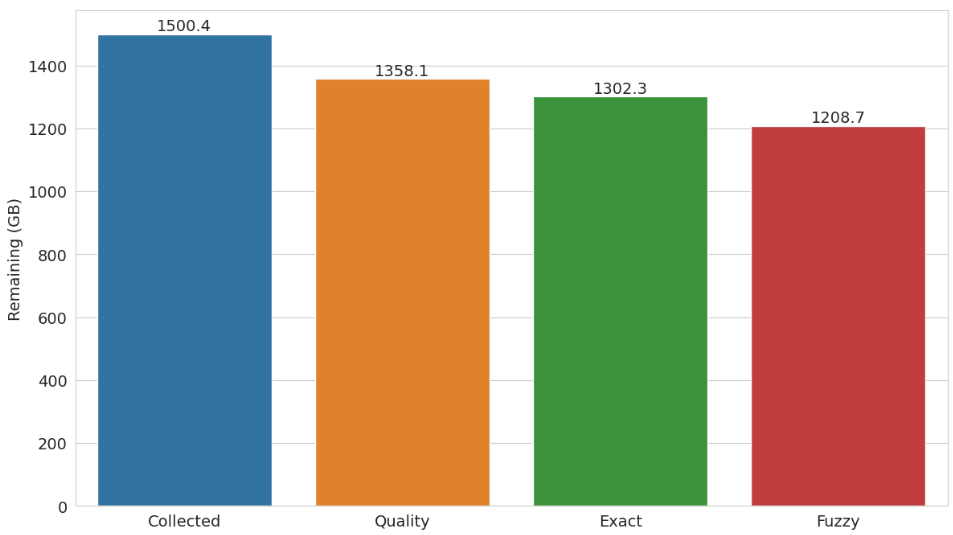}
\centering
\caption{Data quantity remaining after each step in the pipeline.}
\label{fig:remaining_overall}
\end{figure}

Figure \ref{fig:remaining_overall} shows the data size in GB remaining after each step in the data pipeline. This illustrates the importance of all steps, with the most prominent steps being quality-filtering and fuzzy deduplication. In total, almost 300GB (20\%) was removed. 

\begin{figure}[h!]
\includegraphics[width=\linewidth]{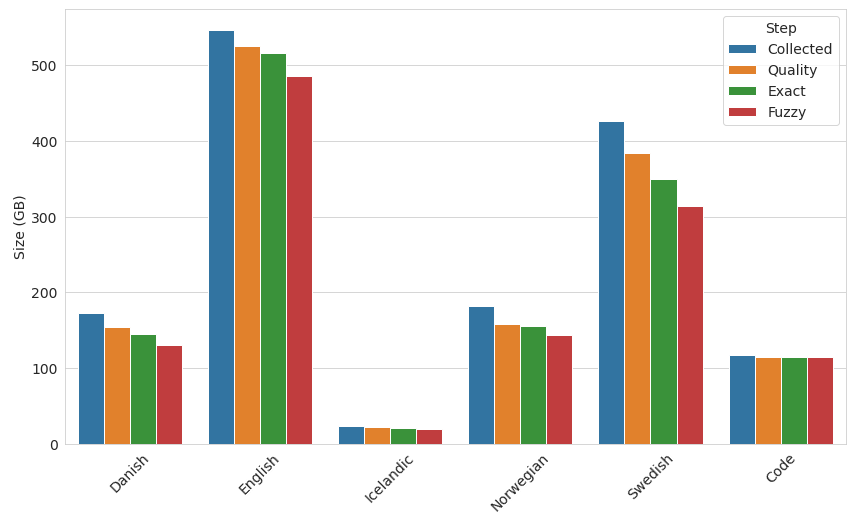}
\centering
\caption{Remaining data for each language after the individual pipeline steps. Note that documents removed by the language filter naturally are not included here, since only languages identified as supported are included.}
\label{fig:lang_multi_bar}
\end{figure}

\begin{figure}[h!]
\includegraphics[width=\linewidth]{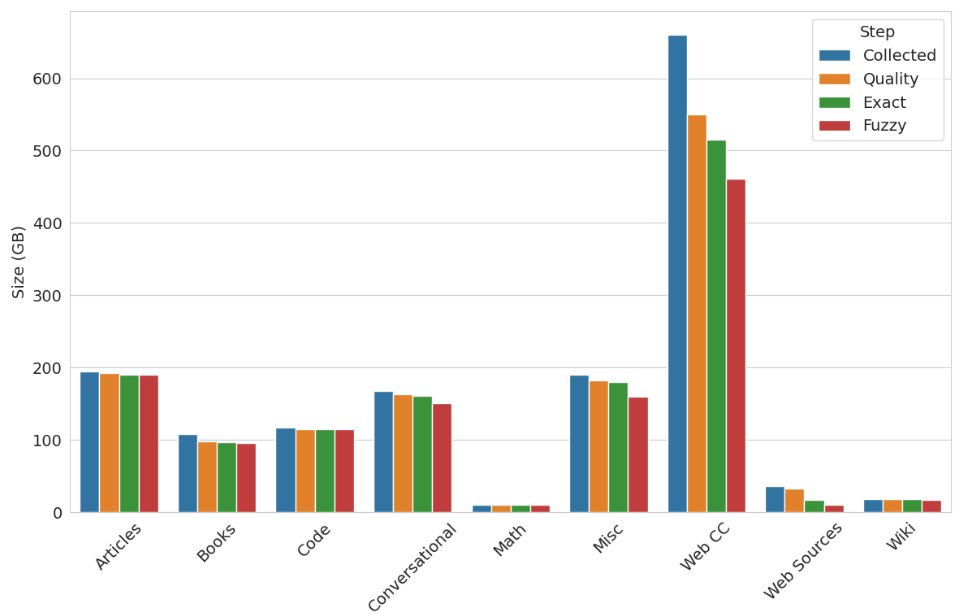}
\centering
\caption{Remaining data in each category after the individual pipeline steps.}
\label{fig:category_multi_bar}
\end{figure}

Figures \ref{fig:lang_multi_bar} and \ref{fig:category_multi_bar} show the same form of information, but for each language and category respectively. The pipeline steps removed similar fractions for each language, with the exception of code where little quality filtering and no deduplication were performed. The same exception for code is seen in the categories. Furthermore, the Web CC category was, by a large margin, affected most by the cleaning process. 

\begin{figure}[h!]
\includegraphics[width=0.7\linewidth]{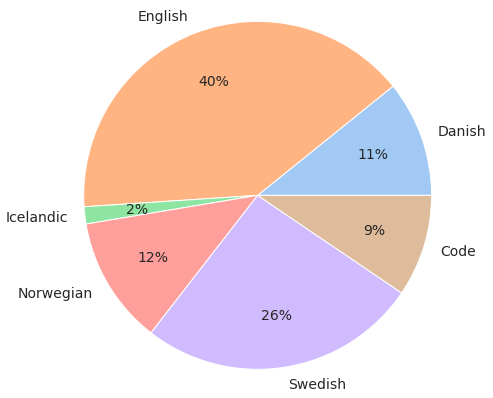}
\centering
\caption{Language distribution of the final dataset. The negligible portion identified as other languages has been omitted. }
\label{fig:lang_distribution}
\end{figure}

\begin{figure}[h!]
\includegraphics[width=0.8\linewidth]{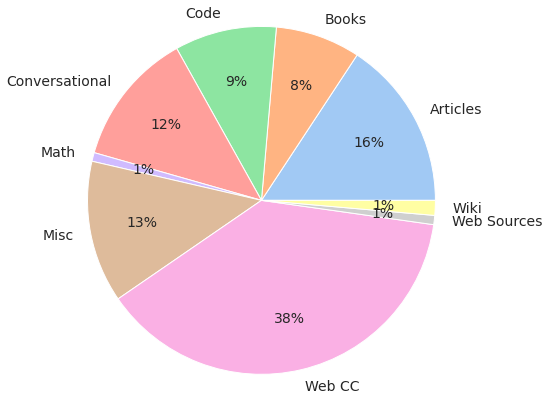}
\centering
\caption{Category distribution of the final dataset.}
\label{fig:category_distribution}
\end{figure}

The pie charts in Figures \ref{fig:lang_distribution} and \ref{fig:category_distribution} give an overview of the language and category compositions of the final dataset. This shows the difficulty of collecting data for low-resource languages such as Icelandic. The largest portion of our dataset consists of English data, and can if desired be adjusted through subset weighting before training. Similarly, Web CC is the most prominent of our categories. Detailed language and category data sizes are shown in Table \ref{fig:category_language_sizes}, with the corresponding fractions in Table \ref{fig:category_language_fractions}. For information about individual sources with their collected and final sizes in GB, the number of documents, and mean document size, see Appendix \ref{sec:appendix_data_sources}.

\begin{table*}[]
\tiny
\begin{center}
\caption{Data sizes for each language and category.}
\label{fig:category_language_sizes}
\begin{tabular}{@{}l|rrrrrrr|r@{}}
\toprule
                        & \textbf{Danish} & \textbf{English} & \textbf{Icelandic} & \textbf{Norwegian} & \textbf{Swedish} & \textbf{Other} & \textbf{Code} & \textbf{Total} \\
\midrule
\textbf{Articles}       & 0.19 GB            & 173.52 GB           & 0 GB                  & 0.01 GB               & 16.49 GB            & 0 GB         &                   & 190.21 GB         \\
\textbf{Books}          & 0.06 GB            & 94.14 GB            & 0 GB                  & 0.04 GB               & 1.15 GB             & 0 GB         &                   & 95.39 GB          \\
\textbf{Conversational} & 2.84 GB            & 81.67 GB            & 0.07 GB               & 0.57 GB               & 65.61 GB            & 0.01 GB      &                   & 150.77 GB         \\
\textbf{Math}           & 0.01 GB            & 4.98 GB             & 0 GB                  & 0.01 GB               & 4.58 GB             & 0.19 GB      &                   & 9.77 GB           \\
\textbf{Miscellaneous}  & 13.85 GB           & 56.31 GB            & 10.26 GB              & 48.48 GB              & 28.85 GB            & 1.8 GB       &                   & 159.55 GB         \\
\textbf{Web CC}         & 111.33 GB          & 60.36 GB            & 8.79 GB               & 90 GB                 & 188.94 GB           & 2.05 GB      &                   & 461.47 GB         \\
\textbf{Web Sources}    & 1.85 GB            & 0.61 GB             & 0 GB                  & 0.03 GB               & 7.83 GB             & 0 GB         &                   & 10.32 GB          \\
\textbf{Wikipedia}      & 0.38 GB            & 14.77 GB            & 0.05 GB               & 0.48 GB               & 1.03 GB             & 0 GB         &                   & 16.71 GB          \\
\textbf{Code}           &                    &                     &                       &                       &                     &              & 114.5 GB          & 114.5 GB          \\
\midrule
\textbf{Total}          & 130.51 GB          & 486.36 GB           & 19.17 GB              & 139.62 GB             & 314.48 GB           & 4.05 GB      & 114.5 GB          & 1208.69 GB        \\
\bottomrule
\end{tabular}
\end{center}
\end{table*}

\begin{table*}[]
\tiny
\begin{center}
\caption{Data fractions in percent for each language and category.}
\label{fig:category_language_fractions}
\begin{tabular}{@{}l|rrrrrrr|r@{}}
\toprule
                        & \textbf{Danish} & \textbf{English} & \textbf{Icelandic} & \textbf{Norwegian} & \textbf{Swedish} & \textbf{Other} & \textbf{Code} & \textbf{Total} \\
\midrule
\textbf{Articles}       & 0.02 \%         & 14.36 \%         & 0.0 \%             & 0.0 \%             & 1.36 \%          & 0.0 \%         &               & 15.74 \%       \\
\textbf{Books}          & 0.0 \%          & 7.79 \%          & 0.0 \%             & 0.0 \%             & 0.1 \%           & 0.0 \%         &               & 7.89 \%        \\
\textbf{Conversational} & 0.23 \%         & 6.76 \%          & 0.01 \%            & 0.05 \%            & 5.43 \%          & 0.0 \%         &               & 12.47 \%       \\
\textbf{Math}           & 0.0 \%          & 0.41 \%          & 0.0 \%             & 0.0 \%             & 0.38 \%          & 0.02 \%        &               & 0.81 \%        \\
\textbf{Miscellaneous}  & 1.15 \%         & 4.66 \%          & 0.85 \%            & 4.01 \%            & 2.39 \%          & 0.15 \%        &               & 13.2 \%        \\
\textbf{Web CC}         & 9.21 \%         & 4.99 \%          & 0.73 \%            & 7.45 \%            & 15.63 \%         & 0.17 \%        &               & 38.18 \%       \\
\textbf{Web Sources}    & 0.15 \%         & 0.05 \%          & 0.0 \%             & 0.0 \%             & 0.65 \%          & 0.0 \%         &               & 0.85 \%        \\
\textbf{Wikipedia}      & 0.03 \%         & 1.22 \%          & 0.0 \%             & 0.04 \%            & 0.09 \%          & 0.0 \%         &               & 1.38 \%        \\
\textbf{Code}           &                 &                  &                    &                    &                  &                & 9.47 \%       & 9.47 \%        \\
\midrule
\textbf{Total}          & 10.8 \%         & 40.24 \%         & 1.59 \%            & 11.55 \%           & 26.02 \%         & 0.34 \%        & 9.47 \%       & 100.0 \%       \\
\bottomrule
\end{tabular}
\end{center}
\end{table*}

\section{Discussion}

Our main guiding principles when collecting the Nordic Pile have been to ensure that the data is both representative and of high quality, while at the same time being of sufficient quantity to enable training of LLMs. To ensure that the models can generalize over a variety of different domains, and that they are useful for a range of different applications, we have collected texts that are representative of a variety of language styles and usages, knowledge domains, and social groups. The mix of editorial sources, such as newspaper articles or texts published on the homepages of governmental authorities and public sector organizations, and user-generated content such as blogs and forums assures a wide range of both styles and topics. In our work with the Nordic Pile, we have made significant efforts to both comply with GDPR and at the same time ensure that the data contains as representative and diverse content as possible.

We have explicitly considered the potential advantages, disadvantages, and risks of including the individual datasets in the Nordic Pile. Advantages and disadvantages relate primarily to the quality of the text sources regarding technical processability. As an example, some datasets primarily contain text extracted from PDFs, but because of the pitfalls of PDF text extraction, such datasets were not included in the Nordic Pile. Another reason for dismissing datasets is the occurrence of computer-generated texts, which was the case e.g.~for a dataset containing subtitles from YouTube. Risks relate primarily to text sources that we know contain, or that are very likely to contain, large amounts of personal information. Such sources have not been included in the dataset. We have, however, {\em not} filtered potentially controversial text content, since that would affect the range of potential applications of a model trained on the data.

Even if we have made considerable efforts to compile the Nordic Pile in a responsible, compliant, and transparent manner, it is clear that the current European legislation does not allow us to distribute and share the processed dataset. This is not a situation we are happy with, but we hope that this paper at least can provide some help for others that undertake their own data collection efforts. We also hope that this paper can help foster a discussion about how we can work with large-scale datasets for LLMs in the Nordic region, in anticipation of future European data initiatives that may facilitate efforts such as this.

In summary, we have created a massive Nordic multilingual dataset consisting of 1.2TB of cleaned and filtered text. The dataset covers the major North Germanic languages Danish, Icelandic, Norwegian, and Swedish, as well as a sizable portion of high-quality English data. The data contains a large variety of genres, domains and topics, and is thoroughly cleaned, filtered and deduplicated in order to ensure that the resulting dataset will provide a high-quality foundation on which to build state-of-the-art Nordic LLMs. In addition to being sufficiently large for training a multi-billion parameter language model, the Nordic Pile dataset also contains a rich typological variety that we hypothesize will be useful for the model’s performance in all of the Nordic languages.


\bibliography{anthology,custom}
\bibliographystyle{acl_natbib}

\clearpage

\appendix

\section{Deduplication Algorithms}
\label{sec:appendix_deduplication_algorithms}

\textbf{Exact Substring Matching} identifies duplicate substrings across documents and removes them from the documents. A naive approach would be quadratic with respect to the number of documents, and the solution used by \citet{deduplicating_2022} uses Suffix Arrays. All documents are concatenated to one massive sequence, for which the Suffix Array is created in linear time. This data structure enables identification of duplicate substrings with linear complexity. This can also be done over tokenized data which may reduce the compute required. \citet{deduplicating_2022} finds and removes all duplicate substrings longer than 50 BPE tokens. 

\textbf{MinHash LSH} identifies approximate duplicates on the document-level, and is widely used in the literature, e.g. \citep{gpt3_2020, pile_2021, gopher_2021}. The core idea of this method is to represent a pair of documents as $C_i$ and $C_j$ as sets $s_i$ and $s_j$, which can be done in many ways \citep{minhashmethods_2016}, and measure their similarity using the Jaccard Similarity (see Equation \ref{eq:jaccard_similarity}). We use this approach for fuzzy deduplication, as it is commonly seen in the literature and has available implementations which we could easily adapt and use\footnote{https://github.com/NVIDIA/Megatron-LM/tree/main/tools/openwebtext}.


\begin{equation}
\label{eq:jaccard_similarity}
    Sim(C_i, C_j) = \dfrac{|s_i \cap s_j|}{|s_i \cup s_j|}    
\end{equation}

A naive approach would measure the similarity of each document pair and would result in a quadratic scaling for the hundreds of millions of documents. This algorithm addresses this and includes three primary steps:

\begin{enumerate}
    \item \textbf{Shingling} converts documents to integer sets through character-based n-grams. We use 10-grams. This step also hashes each shingle to a 32-bit integer, simply to be able to work with integers in the next step, and should not be confused with the hash functions used there.
    
    \item \textbf{MinHashing} converts each large integer set (document) to a small sequence of $p$ integers, that preserves similarity. While this explanation does not provide strong intuition for why the algorithm works, the concrete steps to create these fingerprints are the following:
    
    Create $p$ hash functions, each acting as a permutation operation $\pi$ of the integer set shingles\footnote{http://snap.stanford.edu/class/cs246-2012/slides/03-lsh.pdf}. Define the function $h_\pi(C)$ as the minimum value of the permuted (hashed) shingles sequence $\min \pi(C)$. For each $\pi$, $h_\pi(C)$ converts the shingle sequence to a single integer, resulting in a sequence of $p$ integers for each document, often referred to as fingerprints. We use $p=10$ hash functions, to achieve balance between computational cost and deduplication accuracy.
    
    
    
    
    \item \textbf{Locality-Sensitive Hashing (LSH)} finds signature pairs that are likely created from similar documents. While MinHashing can significantly reduce the computational effort required to find duplicates, by comparing the fingerprints instead of shingles or even whole documents, it still suffers from squared complexity with respect to the number of documents. This step removes the need for comparing all document pairs through hash collisions. 
    
    First, the fingerprints are sliced into $b$ smaller bands. Each band is then independently managed in its corresponding bin, where it can be compared with other documents' band in the same position. A hash-map is created for each bin, mapping a hashed fingerprint slice to a set (bucket) of documents. Whenever multiple bands are hashed to the same bucket, the corresponding documents are duplication candidates. We used $b=2$ bands.
    
    Lastly, all candidates found are iterated and for each candidate pair, the Jaccard Similarity is computed. If the similarity exceeds the Jaccard Threshold the documents are considered duplicates. We used a threshold of 0.5. Now, duplicate documents are connected, and duplicate groups (connected components) can be formed. For each duplicate group, we remove all but one document.
\end{enumerate}


\section{Sharding per language}
\label{sec:shards}

\begin{table}[]
\small
\begin{center}
\caption{Shows for each language, the number of shards, and whether pair-wise inter-shard deduplication was conducted. }
\label{tab:deduplication_shards}
\begin{tabular}{@{}lcc@{}}
\toprule
Language  & Num. Shards & Inter-shard \\ \midrule
Danish    & 4           & Yes         \\
English   & 10          & No          \\
Icelandic & 1           & N/A         \\
Norwegian & 4           & Yes         \\
Swedish   & 9           & Yes         \\ \bottomrule
\end{tabular}
\end{center}
\end{table}

\section{Data Sources}
\label{sec:appendix_data_sources}

A complete overview of the sources of our dataset is illustrated in Table \ref{tab:data_sources}. 

\begin{table*}[]
\tiny
\begin{center}
\caption{Listing of categories and their data sources, with statistics of collected data, and final training data. }
\label{tab:data_sources}
\begin{tabular}{@{}llrrrrr@{}}
\toprule
\textbf{Category}        & \textbf{Source}                & \textbf{Collected Documents} & \textbf{Collected Size} & \textbf{Final Documents} & \textbf{Final Size} & \textbf{Mean Document Size} \\
\midrule
\textbf{Articles}                 & Danish Gigaword                & 0 M                                & 0.4 GB                          & 0 M                            & 0.19 GB                     & 309.24 KB \\
                 & Diva                           & 0.2 M                              & 18.32 GB                        & 0.17 M                         & 16.58 GB                    & 94.87 KB                  \\
                 & The Pile: ArXiv                & 1.26 M                             & 59.52 GB                        & 1.25 M                         & 59.13 GB                    & 47.35 KB                      \\
                 & The Pile: PubMed               & 18.62 M                            & 115.97 GB                       & 18.21 M                        & 114.31 GB                   & 6.28 KB                       \\
\midrule
\textbf{Books}                    
                    & Litteraturbanken               & 0 M                                & 0.31 GB                         & 0 M                            & 0.3 GB                      & 259.58 KB                  \\
                    & The Pile: Books S3             & 0.2 M                              & 107.76 GB                       & 0.17 M                         & 95.08 GB                    & 551.08 KB              \\
\midrule
\textbf{Code}                     & Code Parrot GitHub: JavaScript & 6.39 M                             & 54.22 GB                        & 6.23 M                         & 53.46 GB                    & 8.58 KB      \\
                     & Code Parrot GitHub: Python     & 7.18 M                             & 55.1 GB                         & 7.12 M                         & 54.51 GB                    & 7.66 KB                   \\
                     & Code Parrot GitHub: SQL        & 0.61 M                             & 4.44 GB                         & 0.58 M                         & 3.4 GB                      & 5.85 KB                   \\
                     & Code Parrot GitHub: Shell      & 1.37 M                             & 3.16 GB                         & 1.34 M                         & 3.12 GB                     & 2.34 KB                   \\
\midrule
\textbf{Conversational}           & Anföranden                     & 0.03 M                             & 0.78 GB                         & 0.02 M                         & 0.73 GB                     & 29.64 KB      \\
            & Danish Gigaword                & 0.02 M                             & 1.95 GB                         & 0.02 M                         & 1.58 GB                     & 102.12 KB                           \\
           & Dialog Inpainting              & 11.26 M                            & 24.89 GB                        & 11.16 M                        & 24.75 GB                    & 2.22 KB                             \\
           & Familjeliv                     & 2.68 M                             & 29.93 GB                        & 2.66 M                         & 29.48 GB                    & 11.07 KB                            \\
           & Flashback                      & 3.2 M                              & 37.71 GB                        & 2.92 M                         & 33.75 GB                    & 11.57 KB                            \\
           & Parlai                         & 9.62 M                             & 3.42 GB                         & 1.43 M                         & 1.3 GB                      & 0.91 KB                             \\
           & Reddit                         & 129.39 M                           & 68.92 GB                        & 114.45 M                       & 59.17 GB                    & 0.52 KB                             \\
\midrule
\textbf{Math}                     & English Generated Math         & 56.07 M                            & 5.11 GB                         & 55.36 M                        & 5.07 GB                     & 0.09 KB      \\
                     & Swedish Generated Math         & 56.07 M                            & 5.21 GB                         & 49.67 M                        & 4.71 GB                     & 0.09 KB                    \\
\midrule
\textbf{Miscellaneous}            & DN Summarization               & 0.34 M                             & 1.15 GB                         & 0.28 M                         & 0.92 GB                     & 3.27 KB      \\
            & Icelandic Gigaword             & 4.91 M                             & 10.43 GB                        & 4.09 M                         & 8.85 GB                     & 2.17 KB                             \\
            & Movie Scripts                  & 0 M                                & 0.31 GB                         & 0 M                            & 0.2 GB                      & 131.65 KB                           \\
            & Natural Instructions           & 0.15 M                             & 2.4 GB                          & 0.15 M                         & 2.35 GB                     & 15.85 KB                            \\
            & Norwegian Colossal Corpus      & 20.83 M                            & 43.94 GB                        & 17.28 M                        & 38.65 GB                    & 2.24 KB                             \\
            & OPUS                           & 125.86 M                           & 63.39 GB                        & 125.86 M                       & 63.39 GB                    & 0.5 KB                              \\
            & P3, Public Pool of Prompts     & 58.64 M                            & 33.98 GB                        & 24.59 M                        & 11.67 GB                    & 0.47 KB                             \\
            & The Pile: Stack Exchange       & 15.62 M                            & 34.55 GB                        & 15.25 M                        & 33.53 GB                    & 2.2 KB                              \\
\midrule
\textbf{Web CC}     & LES - Nordic Web Data          & 141.81 M                           & 132.38 GB                       & 79.16 M                        & 76.83 GB                    & 0.97 KB                     \\
                    & MC4                            & 104.79 M                           & 374.67 GB                       & 73.48 M                        & 276.86 GB                   & 3.77 KB                     \\
                   & OSCAR                          & 31.35 M                            & 85.58 GB                        & 19.77 M                        & 49.06 GB                    & 2.48 KB                     \\
                   & The Pile: Open Web Text        & 17.1 M                             & 67.38 GB                        & 14.53 M                        & 58.74 GB                    & 4.04 KB                     \\
\midrule
\textbf{Web Sources}              & Danish Gigaword                & 0.29 M                             & 6.06 GB                         & 0.13 M                         & 1.85 GB                     & 13.92 KB     \\
              & JobTech: Swedish Job Ads       & 6.1 M                              & 12.34 GB                        & 3.61 M                         & 7.02 GB                     & 1.95 KB                             \\
              & Swedish Website Scrape         & 13.68 M                            & 17.22 GB                        & 0.9 M                          & 1.45 GB                     & 1.62 KB                             \\
\midrule
\textbf{Wikipedia}                & Official Wikipedia Dumps       & 22.25 M                            & 17.51 GB                        & 7.64 M                         & 16.71 GB                    & 2.19 KB      \\
\midrule
\textbf{The Nordic Pile} &                                & \textbf{867.89 M}                  & \textbf{1500.41 GB}             & \textbf{659.48 M}              & \textbf{1208.7 GB}          & \textbf{0.55 KB}        \\           
\bottomrule
\end{tabular}
\end{center}
\end{table*}

\section{Quality Filter Configurations}
\label{sec:appendix_filter_configs}

Below follows an enumeration of the quality filters, these filter indices are shown as columns in Table \ref{tab:filter_configs}, illustrating which filters were used for our different subsets. A row either refers to an entire category or one specific data source that required its own configuration.  

\begin{enumerate}
    \item Alpha Present
    \item Blacklist URLs
    \item Digit Fraction
    \item Document Length
    \item Ellipsis To Word Ratio
    \item Flagged Words
    \item Hashtag To Word Ratio
    \item Initial Bullet Point
    \item Mean Line Length
    \item Mean Word Length
    \item Repetitive BSP
    \item Repetitive Gopher
    \item Stop Word
    \item Supported Language
    \item Trailing Ellipsis
\end{enumerate}

So, each data source is mapped to its own configuration if present, otherwise to its category's configuration with the following exceptions/modifications:

\begin{itemize}
    \item The Articles category uses the Books configuration.
    \item The Wikipedia category uses the Web Sources configuration.
    \item Icelandic Gigaword uses the Books configuration.
    \item The Pile: ArXiv uses the stackexchange configuration.
    \item dn\_summarization uses the Books configuration.
    \item movie\_scripts uses the Books configuration.
    \item P3 uses the natural\_instructions configuration.
    \item OPUS uses the Web CC configuration.
\end{itemize}

\begin{table*}[]
\small
\begin{center}
\caption{Binary matrix, depicting the quality filter configurations for individual categories and data sources. Each row corresponds to a data source or category, and each column corresponds to a filter. Element $e_{ij}$ is 1 if filter $j$ was used for data source/category $i$. }
\label{tab:filter_configs}
\begin{tabular}{@{}llllllllllllllll@{}}
\toprule
Data Subset           & \multicolumn{15}{c}{Active Filters}                       \\
\midrule
Books                 & 1 & 0 & 0 & 1 & 1 & 0 & 1 & 1 & 1 & 1 & 1 & 1 & 1 & 1 & 1 \\
Code                  & 0 & 0 & 1 & 1 & 0 & 0 & 0 & 0 & 0 & 0 & 0 & 0 & 0 & 0 & 0 \\
Conversational        & 1 & 0 & 0 & 1 & 1 & 0 & 1 & 1 & 1 & 1 & 0 & 0 & 1 & 1 & 1 \\
Math                  & 0 & 0 & 0 & 0 & 1 & 0 & 1 & 1 & 0 & 0 & 0 & 0 & 0 & 0 & 1 \\
Web CC                & 1 & 1 & 0 & 1 & 1 & 1 & 1 & 1 & 1 & 1 & 1 & 1 & 1 & 1 & 1 \\
Web Sources           & 1 & 0 & 0 & 1 & 1 & 0 & 1 & 1 & 1 & 1 & 1 & 1 & 1 & 1 & 1 \\
natural\_instructions & 0 & 0 & 0 & 1 & 1 & 0 & 1 & 1 & 1 & 1 & 0 & 0 & 0 & 0 & 1 \\
ncc                   & 1 & 0 & 0 & 1 & 1 & 0 & 1 & 1 & 1 & 1 & 1 & 1 & 1 & 1 & 1 \\
pubmed\_central       & 0 & 0 & 0 & 1 & 1 & 0 & 0 & 1 & 1 & 0 & 0 & 0 & 1 & 1 & 1 \\
stackexchange         & 0 & 0 & 0 & 1 & 1 & 0 & 0 & 1 & 0 & 0 & 0 & 0 & 0 & 1 & 1 \\
\bottomrule
\end{tabular}
\end{center}
\end{table*}

\section{Fuzzy Duplicate Group Examples}
\label{sec:appendix_fuzzy_groups}

Table \ref{tab:fuzzy_groups} illustrates some examples of documents that are removed in the fuzzy deduplication step and shows typical documents that are similar but not identical.

\begin{table*}[]
\tiny
\begin{center}
\caption{Examples of fuzzy duplicate group documents. Group size defines the number of documents that are considered duplicates of each other within the group. For each example document, a snippet of the first 200 characters is shown. Line breaks have been omitted.}
\label{tab:fuzzy_groups}
\begin{tabular}{@{}lp{0.7\linewidth}@{}}
\toprule
\textbf{Group Size}           & \textbf{Example Document Snippets}\\
\toprule
\textbf{1023616}     &
\textit{Ange sannolikheten att välja 3 g då tre bokstäver väljs utan återsättning från gggjjgggg.
Svaret är: 5/12} \\
\midrule \\
             &
\textit{Vad är sannolikheten att välja 2 q and 2 j då fyra bokstäver väljs utan återsättning från \{j: 2, h: 1, q: 7\}?
1/10} \\
\midrule \\
             &
\textit{Hitta sannolikheten att sekvensen är zz då två bokstäver väljs utan återsättning från cgzcgcxxhggggxc.
Svaret är 0} \\
\bottomrule
\textbf{1006}    & 
\textit{Given During the 2008–09 season AFC Ajax participated in the Dutch Eredivisie, the KNVB Cup and the UEFA Cup. The first training took place on Monday July 14, 2008. The traditional AFC Ajax Open Day was on Tuesday August 5, 2008, followed by a testimonial match for the retired former Ajax defender Jaap Stam. Is it guaranteed true that "Jaap Stam retired from AFC Ajax because he wanted to pursue an} \\
\midrule \\
        &
\textit{Given During the 2008–09 season AFC Ajax participated in the Dutch Eredivisie, the KNVB Cup and the UEFA Cup. The first training took place on Monday July 14, 2008. The traditional AFC Ajax Open Day was on Tuesday August 5, 2008, followed by a testimonial match for the retired former Ajax defender Jaap Stam. Is it guaranteed true that "Jaap Stam only played defender his whole career. "? Yes, no, o} \\
\midrule \\
        &
\textit{During the 2008–09 season AFC Ajax participated in the Dutch Eredivisie, the KNVB Cup and the UEFA Cup. The first training took place on Monday July 14, 2008. The traditional AFC Ajax Open Day was on Tuesday August 5, 2008, followed by a testimonial match for the retired former Ajax defender Jaap Stam. Keeping in mind the above text, consider: Jaap Stam will come out of retirement and play profe} \\
\bottomrule
\textbf{638}     &
\textit{PUBLICERADES 30 okt 2014 11:06
Stormen Ivar
Detta är en arkiverad sida över händelserna efter den 12 december då stormen Ivar drog in över delar av norra Sverige. Sidan uppdateras inte längre.
PUBLICERADES 3 nov 2014 14:38
Stormen Simone
Detta är en arkiverad sida med information om stormen Simone som drabbade delar av södra Sverige den 27-29 oktober 2013. Sidan uppdateras inte längre.
PUBLICERADE} \\
\midrule \\
        &
\textit{PUBLICERADES 5 nov 2014 13:35
Stormen Per
Två år efter stormen Gudrun drabbades elbolag och skogsägare i södra och västra Sverige än en gång av en svår storm. Ett intensivt lågtryck bildades den 13 januari strax väster...
PUBLICERADES 3 nov 2014 14:38
Stormen Simone
Detta är en arkiverad sida med information om stormen Simone som drabbade delar av södra Sverige den 27-29 oktober 2013. Sidan uppdat} \\
\bottomrule
\textbf{583}        &
\textit{Säljare Vitvaror
Vi söker drivna och ambitiösa säljare tillVitvaror Arbetsbeskrivning I rollen som Butikssäljare på Media Markt arbetar du huvudsakligen på någon av våra nio olika säljavdelningar: Data, Foto, HiFi, Spel, Film \& Musik, Telefoni, Tillbehör, TV, Stora Vitvaror samt Små Vitvaror.Vid behov kommer du vara behjälplig på övriga avdelningar och expediera kunder.På Media Markt arbetar vi in} \\
\midrule \\
        &
\textit{Helgsäljare till TV-avdelningen i Heron City
Vi växer och söker därför en driven och ambitiös säljare till en av Sveriges största TV-avdelningar Arbetsbeskrivning I rollen som säljare på Media Markt arbetar du huvudsakligen på någon av våra nio olika säljavdelningar: Data, Foto, Ljud, Spel, Film \& Musik, Telefoni, Tillbehör, TV, Stora Vitvaror samt Små Vitvaror.Vid behov kommer du vara behjälplig} \\
\bottomrule
\textbf{264}     &
\textit{Extra- och deltidsmedarbetare sökes 
XXL - ett eldorado för sport- och vildmarksälskare. Våra stora varuhus erbjuder ett enormt utbud av kända varumärken till extra låga priser inom sport- och vildmar} \\
\midrule \\
        &
\textit{Säljare  2:e man skidor \& cykel
XXL - ett eldorado för sport- och vildmarksälskare. Våra stora varuhus erbjuder ett enormt utbud av kända varumärken till extra låga priser inom sport- och vildmarkspro} \\
\midrule \\
        &
\textit{Säljare Sportkläder
XXL - ett eldorado för sport- och vildmarksälskare. Våra stora varuhus erbjuder ett enormt utbud av kända varumärken till extra låga priser inom sport- och vildmarksprodukter. Hos } \\
\bottomrule
\end{tabular}
\end{center}
\end{table*}

\section{Data Pipeline Deviations}
\label{sec:appendix_deviations}

\begin{itemize}
    \item The Familjeliv data was included after the deduplication step and was, therefore, deduplicated in isolation. We believe the impact of this is minimal since this data was manually scraped by us and should not be present in other datasets using the same conversational format.
    \item The OPUS data which is composed of parallel sentences was cleaned and deduplicated in isolation prior to being formulated with prompt templates. 
\end{itemize}

\end{document}